\documentclass[twoside]{article}

\usepackage{aistats2019}

\usepackage[round]{natbib}

\usepackage{url}
\usepackage[utf8]{inputenc}
\usepackage{graphicx}
\usepackage{amsmath,amssymb}
\usepackage{sansmath}
\usepackage{multirow, array}
\usepackage{booktabs}
\graphicspath{ {figures/} }
\usepackage{graphics}
\usepackage{tikz}
\usepackage{graphicx}
\usepackage[export]{adjustbox}
\usetikzlibrary{positioning}
\usepackage{pgfplots}

\pgfplotsset{
    compat=1.11, %
    width=9.5cm,
    height=5cm,
}
\pgfkeys{/pgfplots/scale/.style={
  x post scale=#1,
  y post scale=#1,
  z post scale=#1}
}
\pgfkeys{/pgfplots/axis labels at tip/.style={
    xlabel style={at={(current axis.right of origin)}, xshift=1.5ex, anchor=center},
    ylabel style={at={(current axis.above origin)}, yshift=1.5ex, anchor=center}}
}

\usepackage{color}
\definecolor{atomictangerine}{rgb}{0.8, 0.2, 0.1}
\definecolor{turq}{rgb}{0.0, 0.5, 0.5}
\definecolor{darkturq}{rgb}{0.0, 0.25, 0.25}
\definecolor{bright}{rgb}{0.8, 0.1, 0}
\definecolor{darkgray}{gray}{0.3}
\definecolor{mahogany}{rgb}{0.6, 0.05, 0.05}
\definecolor{pink}{rgb}{1,0.05,0.6}

\usepackage{algorithm}
\usepackage{algorithmic}
\begin{document}
\twocolumn[
    \aistatstitle{Few-Shot Learning with Per-Sample Rich Supervision}
    \aistatsauthor{Roman Visotsky \And Yuval Atzmon \And Gal Chechik}
    \aistatsaddress{Bar-Ilan University \And Bar-Ilan University \And Bar-Ilan University\\ NVIDIA } 
]

\begin{abstract}
Learning with few samples is a major challenge for parameter-rich models like deep networks. In contrast, people learn complex new concepts even from very few examples, suggesting that the sample complexity of learning can often be reduced. Many approaches to few-shot learning build on transferring a representation from well-sampled classes, or using meta learning to favor architectures that can learn with few samples.  Unfortunately, such approaches often struggle when learning in an online way or with non-stationary data streams. 

Here we describe a new approach to learn with fewer samples, by using additional information that is provided per sample. Specifically, we show how the sample complexity can be reduced by providing semantic information about the relevance of features per sample, like information about the presence of objects in a scene or confidence of detecting attributes in an image. We provide an improved generalization error bound for this case. We cast the problem of using per-sample feature relevance by using a new ellipsoid-margin loss, and develop an online algorithm that minimizes this loss effectively. Empirical evaluation on two machine vision benchmarks for scene classification and fine-grain bird classification demonstrate the benefits of this approach for few-shot learning.

\end{abstract}

\newcommand{\reals}{\mathbb{R}}
\newcommand{\D}{{\Sigma}}
\newcommand{\di}{{S_i}}
\newcommand{\diinv}{{S_i^{-1}}}
\newcommand{\qx}{{\mathbf{q}}}
\newcommand{\aw}{{\mathbf{a}}}
\newcommand{\w}{{\mathbf{w}}}
\newcommand{\vvec}{{\mathbf{v}}}
\newcommand{\x}{{\mathbf{x}}}
\newcommand{\z}{{\mathbf{z}}}
\newcommand{\uu}{{\mathbf{u}}}
\newcommand{\bb}{{\mathbf{b}}}
\newcommand{\around}{{\hat{{\x}}}}
\newcommand{\cc}{{\x}}
\newcommand{\scaled}{{\mathbf{u}_i}}

\newcommand{\Si}{{S_i}}
\newcommand{\Siinv}{{S_i^{-1}}}
\newcommand{\Sinv}{{S^{-1}}}
\newcommand{\Sinvi}{{S_i^{-1}}}
\newcommand{\SinviT}{{S_i^{-1}}^T}
\newcommand{\xxi}{{\mathbf{\x_{i}}}}
\renewcommand\vec[1]{\mathbf{#1}}

\newcommand{\ltwonorm}[1]{\left\lVert#1\right\rVert_{2}}
\newcommand{\snorm}[1]{\left\lVert#1\right\rVert_{\Sinvi}}
\newcommand{\snormsq}[1]{\left\lVert#1\right\rVert_{\Sinvi}^2}

\newcommand{\wneg}{{\w_{neg}}}
\newcommand{\wpos}{{\w_{pos}}}
\newcommand{\vneg}{{\vvec_{neg}}}
\newcommand{\vpos}{{\vvec_{pos}}}
\newcommand{\wnegt}{{\w_{neg}^{new}}}
\newcommand{\wpost}{{\w_{pos}^{new}}}
\newcommand{\wnegto}{{\w_{neg}}}
\newcommand{\wposto}{{\w_{pos}}}
\newcommand{\vnegt}{{\vvec_{neg}^{new}}}
\newcommand{\vpost}{{\vvec_{pos}^{new}}}
\newcommand{\vnegto}{{\vvec_{neg}}}
\newcommand{\vposto}{{\vvec_{pos}}}
\newcommand{\loss}{\text{loss}_{El}}

\newcommand\ignore[1]{}
\renewcommand\eqref[1]{{Eq.~(\ref{#1})}}
\newcommand\figref[1]{{Fig.~(\ref{#1})}}
\newcommand{\argmax}{\mathop\text{argmax}}

\section{Introduction}

People can learn to recognize new classes from a handful of examples. In contrast, deep networks need large  labeled datasets to match human performance in object recognition, and perform poorly unless the  data covers well the distribution of samples per class. This performance gap suggests that there are fundamental factors that could reduce the samples complexity of existing learning system. 

\textit{Few-shot learning} (FSL) becomes a real challenge in many domains where it is hard to collect many labeled samples per-class. For example, in fine-grained object recognition, the number of classes may be extremely large, and because the distribution of classes in nature is highly unbalanced, tail-concepts typically have only few samples. As a second  important case, in numerous learning applications, the data is non-stationary and classifiers have to learn in an online way. In this settings, they repeatedly suffer a cost for every wrong decision, and therefore have to quickly adapt based on a few samples. 

Many approaches to few-shot learning and zero-shot learning (ZSL) are based on learning a  representation using well-sampled classes and then using that representation to learn new classes with few samples \citep{snell2017prototypical,vinyals2016matching,hariharan2016low,LAGO,COSMO}. In a second line of approaches, \textit{meta-learning}, a learner is trained to find an inductive bias over the set of model architectures that benefits FSL \citep{ravi2016optimization,finn2017model}. 
Unfortunately, these approaches may not be feasible in the online learning setup where a cost is incurred for every prediction made.

The current paper therefore proposes a complementary approach, inspired by how people learn faster by integrating side information about samples, classes and features. Specifically, when people learn new concepts from few labeled samples ${x_i, y_i}$, they can also effectively use additional per-sample information $z_i$ that provides an inductive bias about the model to be learned. Broadly speaking, such \textit{rich supervision} (RS) may appear in many flavors. For example, classes can be accompanied by their definition, samples can be accompanied by an ``explanation'' of classification \citep{thrun2012explanation,Aodha_2018_CVPR,su2017interpretable}, and features may have names, which can provide priors about their semantics \citep{parikh2011interactively}. Explaining or describing a class with natural language is a hallmark of human learning, allowing people to quickly understand very complex and abstract classes when taught by a human teacher \citep{elhoseiny2017link}. Learning with rich human supervision has two address two major challenges. First, one has to collect rich supervision from human raters, which may be hard to scale. Second, one needs to find ways to integrate the rich supervision into the model effectively. 

Here we focus on a specific type of rich supervision and address these two challenges. We study the case where each sample is accompanied with information about features that are relevant for classification in a sample. More specifically, we study a learning architecture where classification is based on intermediate representation with named entities, like attributes or detected objects. In this setup, we show that it is possible to use open-world tags provided by raters, by mapping them to the intermediate entities. This approach also addresses the second challenge, collecting data at scale, because it is cheap to collect sparse information about features at scale \citep{branson2010visual}. For instance, when human raters provide ground truth labeling of an image, they can easily provide text tags explaining or justifying their decision. We demonstrate below two different datasets where such information is available, and show how the text tags can be mapped onto an internal network representation. 

We formulate the problem in the context of online learning. We design a new,  ellipsoid-margin, loss that takes into account the side-information available, and describe an online algorithm for minimizing that loss efficiently. We test the algorithm with two datasets and find that improves over existing baselines: The SUN benchmark dataset for visual scene classification where objects occurrence are used for RS and the CUB bird image benchmark dataset where attributes are used for RS. 

The novel contributions of this paper are as follows. 
(1) First, we describe the general setup of learning with per-sample side-information and discuss the special case of learning with per-sample information about feature uncertainty and relevance as a special case. (2) We prove a theorem showing how per-sample information can reduce the sample complexity. (3) We then describe Ellipsotron, an online learning algorithm for learning with rich supervision about feature uncertainty, which efficiently uses per-class and per-sample information. (4) We demonstrate empirically how rich supervision can help even in a case of strong-transfer, where expert feedback is provided in an unrestricted language, and that feedback is transferred {\em without learning} to a pretrained internal representation of the network. (5) Finally, we demonstrate the benefit of empirical supervision at the sample level and class level on two standard benchmarks for scene classification and fine-grained classification.

\section{Related Work}
\textbf{Few-shot learning.}
FSL gained significant interest in the past few years, and the relevant literature is extensive. We refer the reader to recent relevant review on zero-shot learning \citep{xian2017zero}, and list very partial recent literature here. A main thrust in FSL focuses on transferring a learned representation from rich-sampled classes to classes with fewer samples. \citet{hariharan2016low,vinyals2016matching}. 
FSL also benefits from {\em meta learning}, which can be used to find architectures where FSL is most effective \citep{ravi2016optimization,finn2017model}. Meta learning assumes that it is possible to repeatedly sample from the distribution of tasks. Our current work operates under the online model were each prediction made also suffers a cost, hence meta learning is not applicable. 

\textbf{Learning with feature feedback.}
A special case of rich supervised learning, when an expert provide explicit feedback about feature importance. Several authors studied this regime for batch learning.  \citet{druck2007reducing} incorporated user-provided information about feature-label relations into the objective. \citet{zaidan2007using} introduced {\em authors rationales}, where a human expert provides hints about feature relevance for classification. The rationales were then used to build “contrasts examples”, by masking out irrelevant features, and these were used for adding ranking-loss components to the objective. Their experiments used dense annotation from raters about movie ratings. A similar approach is taken in \citep{sun2005explanation,sun2006explanation,sun2007robustness}. \citet{small2011constrained} augmented SVM using a set of order constraints over weights, for example, by having an expert provide the learning with the information that the weight of feature $i$ should be larger than that of feature~$j$. \citet{chechik2007max} described a large-margin approach for learning with structurally missing features, which is related to the current paper. \citet{branson2010visual} describes a method to recognize bird species with a human-in-the-loop at inference time,  where question selection is assisted by a machine vision system. \citet{raghavan2006active} studied an active learning setup where raters are asked about relevance of features. \cite{Aodha_2018_CVPR,su2017interpretable} described learning with feedback that highlights informative areas in an image.

Most relevant to this paper is the recent work by \citet{poulis2017learning}. They studied the case of user-provided feedback about features for binary classification and proposed two types of algorithms. First, less relevant to this work, a two-stage probabilistic model where features (terms) are mapped into topics, and these in turn determine the label in a disjunctive way. Second, an SVM applied to rescaled features (Algorithm 4, SVM-FF). The latter is related to our Ellipsotron but differs in the following important ways. First, the fundamental difference is that SVM-FF rescales all data using s single shared matrix, while {\em our approach is class specific or sample specific}. Second, we present an online algorithm. Third, our loss is different, in that samples are only scaled to determine if the margin constrain is obeyed, but the loss is taken over non-scaled samples. This is important since rescaling samples with different matrices may change the geometry of the problem. Indeed, our evaluation of an online-variant of SVM-FF performed poorly (see Eq. 7 below) compared to the approach proposed here.
Also very relevant, is the work of \cite{Dasgupta_NIPS2018_7651}. They address learning a multi-class classifier using simple explanations they call "discriminative features", and analyze learning a particular subclass of DNF formulas with such explanations.

\section{Rich Supervised Learning}
The current paper considers using information about features per-sample. It is worth viewing it in the  more general context of rich supervision (RS). As in standard supervised learning, in RS we are given a set of $n$ labeled samples $(x\in {\cal{X}}, y \in \cal{Y})$ drawn from a distribution $D$, and the goal is to find a mapping $f_W: X \rightarrow Y$ to minimize the expected loss $E_D[loss(f_W(x_i),y_i)]$. In RS, at training time, each labeled sample is also provided with an additional side information $z\in\cal{Z}$. Importantly, $z$ is not available at test time, and only used as an inductive bias when training $f$.

Rich supervision can have many forms. It can be about a class (hence $z_i = z_j$ iff $y_i=y_j$ $\forall$ samples $i,j$), as with a class description in natural language ("zebras have stripes"). It can be about a sample, e.g., providing information about the uncertainty of a label $y_i$, the important of a  sample $\x_i$ ("this sample is important"), or the importance of features per sample ("look here") . 
Finally, it can also be provided as feedback about a set of detectors applied to a sample, $z_i \in f_W(x_i)$. This is the case studied in this paper, where we map images to a predefined set of detectors, characterized by natural language terms.

A key component of learning with rich supervision is to obtain a rich signal $z_i$ that contains sufficient inductive bias to improve training. In some cases experts can provide direct feedback about raw features of the problem. For example, a radiologist interpreting an X-ray scan may mark certain areas in a scan to highlight their importance. In other cases, feedback from experts is not directly about sample features, but about a high level representation. For example, a bird enthusiast recognizes a bird by its long bill or red crown. In these cases, one has to map the feedback into an internal representation. We show below how it is possible to use feedback as free form tags without the expert being familiar with the internal representation of the model.

How can RS help? First, it could change the optimal solution of the problem, by changing the loss  \citep[as in][]{poulis2017learning}, or by changing the training set, which in turn changes the optimal classifier \citep[as in][]{zaidan2007using}. RS could also speed up the learning process (fewer samples) by providing the learner with information about the geometry of the error space, e.g. through information about the gradient at some region of the search space. 

\section{learning with Per-Sample Feature Information}
We focus here on the online learning setting for multiclass classification. An online learner repeatedly receives a labeled sample $\x_i$ (with $i=1,\ldots,n$), makes a prediction $\hat{y_i} = f_W(x_i)$ and suffers a cost $loss(\w; \x_i, y_i)$. We explore a specific type of rich supervision $z_i$, providing feedback about features per samples. Specifically, in many cases it is easy to collect information from raters about high-level features being present and important in an image. For instance, raters could easily point that an image of a bathroom contains a sink, and that side information can be added to pre-trained detectors of a sink in images.

{\bf The key technical idea} behind our approach is to define a sample-dependent margin for each sample, whose multidimensional shape is based on the known information about the uncertainty about features for that particular sample. Importantly, this is fundamentally different from scaling the samples. This point is often overlooked because when all samples share the same uncertainty structure, the two approaches become equivalent. Unfortunately, when each sample has its own multidimensional uncertainty scale, scaling samples might completely change the geometry of the problem. We show how to avoid this issue by rescaling the margins, rather than the data samples. 

\begin{figure}[ht]
    \centering
    \includegraphics[trim=10 40 100 50,clip,width=0.7\linewidth]{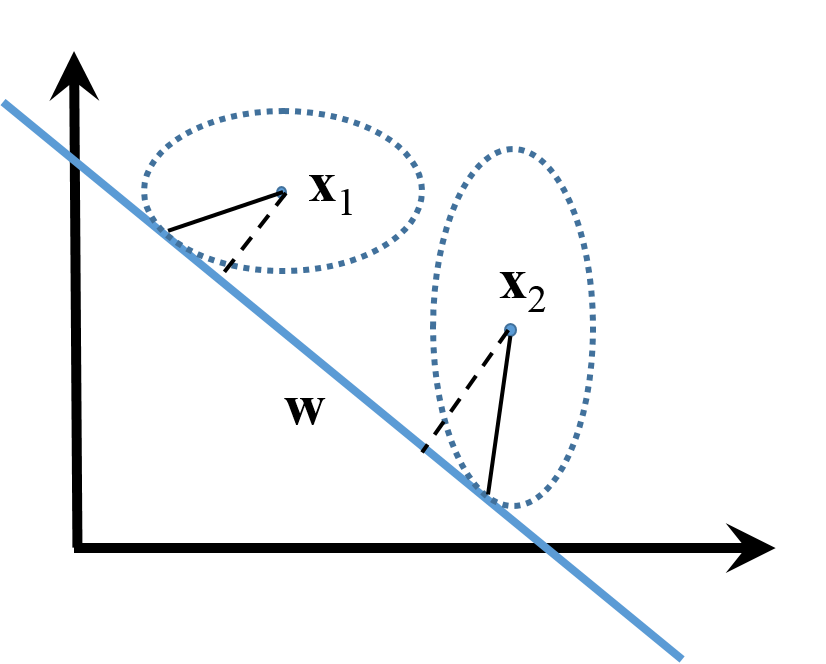}
    \vspace{-10pt}
    \caption{\textit{Illustration of ellipsoid margin per sample. Each sample $\x_i$ has its own uncertainty ellipse (dotted lines) parametrized by a matrix $S_i$, and inducing the ellipsoid margin, see \eqref{eq:binary-ellipsoid-loss}. For a spherical ellipsoid, the margin becomes equal to the standard margin (dashed lines).}\newline}
    \label{fig:ellipsoid}
\end{figure}

To take this information into account, we treat each sample $i$ as if it is surrounded by an ellipsoid centered at a location $\x_i$, see \figref{fig:ellipsoid}. The ellipsoid is parametrized by an \textit{``uncertainty matrix''} $z_i=\di \in \reals_+^{d\times d}$, and represents the set of points where a sample might have been if there was no measurement noise.
It can also be thought as reflecting a known noise covariance around a measurement. When that covariance has independent features, $\di$ is a positive diagonal matrix $\di=diag(s_1,...,s_d)$ ($s\! >\! 0, j=1,\ldots,d$) that represents the uncertainty in each dimension of the sample $\x_i$. In this case, linearly transforming the space using  $\diinv$ makes the uncertainty ellipsoid $S_i$ a symmetric sphere. When the matrix $\diinv$ is diagonal, it can be interpreted as capturing a measure of confidence or precision over the features of the sample $x_i$

We first define an ellipsoid loss for the binary classification case and later extend it to the multiclass case:
\begin{equation}
   loss(\w; \cc, y) = 
   \begin{cases}
        0   &\,\,\,\,\,\min\limits_{ \around \in \cal{X}_S} \,\,  y\w^T\around>0\\    
        1-y\w^T\cc \,\, &\text{ otherwise} \quad
    \label{eq:binary-ellipsoid-loss}
    \end{cases}
\end{equation}
where $\cal{X}_S = {\left\{\around : ||\around - \cc||_{S^{-1}} \le 1/||\w||_S \right\}}$ and $||\x||_S^2 = \x^T S^T S\x$ is the Mahalanobis norm corresponding to the matrix $S^TS$, hence minimization is over the set of points $\around$ that are ''inside",  or $S$-close to, the centroid $\cc$. Intuitively, the conditions in the loss mean that if all points inside the ellipsoid are correctly classified, no loss is incurred.

This definition of the loss extends the standard margin hinge loss, since when $S$ is the identity matrix, the following holds 
\begin{eqnarray*}
    \min_{||\around-\cc|| \le 1/||\w||} y w^T \hat{\x} &=& \min_{||\uu|| \le 1/||\w||} y \w^T (\x+\uu)  \\ \nonumber
    &=& y \w^T \x + \min_{||\uu|| \le 1/||\w||} y \w^T\uu \quad.
\end{eqnarray*} 
The second term is minimized when $\uu = -\w/||w||$, yielding $y \w^T \x - 1$., hence the loss of \eqref{eq:binary-ellipsoid-loss} becomes equivalent to the standard margin loss 
\begin{equation}
   loss(\w; \cc, y) = 
   \begin{cases}
        0  & y\w^T\cc > 1\\    
        1-y\w^T\cc & \text{otherwise}\, .  
\end{cases}
\end{equation}
For the multiclass case, we follow \cite{crammer2006online} and consider a weight vector that is the difference between the positive class and the hardest negative class $\Delta \w = \wpos-\wneg$. We define the multiclass Ellipsoid loss
\begin{equation} \label{eq:multiclass-ellipsoid-loss}
        \loss(W; \cc, y) \!\!= \!\!\\
        \begin{cases}
            0 & \!\!\min\limits_{\around \in \cal{X}_S}\Delta\w^T\around\!>\!0\\    
            1-\Delta\w^T\cc \quad &\text{otherwise}\,.
    \end{cases}
\end{equation}
Assuming that $||\Delta\w||>0$, this loss also becomes equivalent to the standard hinge loss for the identity matrix  $S=I$.

\section{Ellipsotron}
We now describe an online algorithm for learning with the loss of \eqref{eq:multiclass-ellipsoid-loss}. Since it is hard to tune hyper parameters when only few samples are available, we chose here to build on the passive-aggressive algorithm \citep[PA,][]{crammer2006online}, which is generally less sensitive to tuning the aggressiveness hyper parameter.
Our approach is also related to the Ballseptron \citep{shalev2005new}. 

The idea is to transform each sample to a space where it is spherical, then apply PA updates in the sample-dependent scaled space. More formally, for each sample, the algorithm solves the following optimization problem:
\begin{flalign}
    \label{eq:ellipsotron-optim}
    \min_W &\snormsq{W - W^{t-1}}  + C \,\, \loss(W; \cc, y).
\end{flalign}
Similar to PA, it searches for a set of weights $W$ that minimize the loss, while keeping the weights close to the previous  $W^{t-1}$. Different from PA, the metric it uses over $W$ is through the $S$ matrix of the current example. This reflects the fact that similarity over W should take into account which features are more relevant for the sample $\x$.

{\bf Proposition:} {\em The solution to \eqref{eq:ellipsotron-optim} is obtained by the following update steps:}
\begin{equation}
    \begin{split}\label{eq:W_update}
        & {(\wpost)}^T \leftarrow \wposto^T + \frac{\loss}{2||\diinv \x_i||^2 + \frac{1}{2C}} \diinv^T \diinv \x_i \\
        & {(\wnegt)}^T \leftarrow \wnegto^T - \frac{\loss}{2||\diinv \x_i||^2 + \frac{1}{2C}} \diinv^T \diinv \x_i \quad.
    \end{split}
\end{equation}

The full proof is given in the appendix.

\begin{algorithm}[t]
   \caption{}
   \label{alg}
\begin{algorithmic}[1]
   \STATE {\bfseries inputs:} A dataset $\{x_i, y_i, S_i\}_{i=1}^{N}$ of samples, labels and rich-supervision about feature uncertainty;
   A constant $C$
   \STATE {\bfseries initialize:} Set  $W  \leftarrow 0$
   \FOR{\textbf{each} sample $i \in [1 \dots N]$}
   \STATE Set $pos \leftarrow y_i$; true label weights column index.
   \STATE Set $neg \leftarrow \argmax\limits_{n \neq y_i} \w_n^T\x_i $; the hardest false label $n$ for classifying $\x_i$ by $\w_n$, weights column index.
   \STATE Update the columns $pos$, $neg$ of $W$ using \eqref{eq:W_update}
   \ENDFOR
\end{algorithmic}
\end{algorithm}

\section{Generalization Error bound}
We prove a generalization bound for learning linear classifiers from a hypothesis family $\mathcal{F}$ for a set of $n$ i.i.d. labeled samples $(\x_i, y_i)$. Each sample has its own uncertainty matrix $\Sinvi$.
Let $\mathcal{L}$ be the empirical loss $\hat{\mathcal{L}} = \sum_{i=1} loss(\w^T\x_i, y_i)$, and the true loss $\mathcal{L} = E_{p(x,y)} loss(\w^T\x_i, y_i)$, the following relation holds: 

\paragraph{Theorem: }
For a loss function that is upper bounded by a constant $M_l$ and is Lipschitz in its first argument. For the family of linear separators 
$\mathcal{F} = {\w : \sum_i||\w||_\Sinvi \le \sum_i||\w^*||_\Sinvi}$, and for any positive $\delta$ we have with probability $\ge 1-\delta$ and $\forall f\in\mathcal{F}$:
\begin{eqnarray}
     {\mathcal{L}}(f) &\!\!\le\!\!&
    {\hat{\mathcal{L}}}(f) + 2 ||\w^*||_2  \max_{\x_i \in \mathcal{X}} \sqrt{||x_i||_{\Sinvi}} \sqrt{\frac{2}{n}} \\ \nonumber
    & & + M_l\sqrt{\frac{1}{2n}\log(\frac{1}{\delta})}\quad,
\end{eqnarray}
where $w^*$ is a target classifier, and the max goes over the space of samples, with each sample having its predefined corresponding uncertainty matrix $\Siinv$.

The meaning of this theorem is as follows. Consider a case where some dimensions of $\x_i$ are more variable, for example if contaminated with noise that has different variance across different features. The uncertainty matrix $\Sinvi$ matches the dimensions of $\x_i$ such that higher-variance dimensions correspond to smaller magnitude $\Sinvi$ entries. In this case,
$\snorm{\x_i} < \ltwonorm{\x_i}$ hence the theorem leads to a tighter generalization bound, reducing the sample complexity.
As a specific example, for a diagonal $\Sinvi$ with only $k$ non-zero values on the diagonal, {\bf the effective dimension of the data is reduced from $d$ to $k$, even if these $k$ values vary from one sample to another}. This can dramatically reduce sample complexity in practice, because very often, even if a dataset occupies a high dimensional manifold, only a handful of features are sufficient for classifying each sample, and these features vary from one sample to another.

\paragraph{Proof:}
The proof is based on a proof for the case of a single confidence matrix by \cite{poulis2017learning}. First, we use a result by Bartlett and Mendelson (2003)
\begin{equation}
    \forall f\in\mathcal{F} \mathcal{L}(f) \le 
    \hat{\mathcal{L}}(f) + 2 \mathcal{R}_n + M_l\sqrt{\frac{1}{2n}\log(\frac{1}{\delta)}}
\end{equation}
where $\mathcal{R}(f)_n$ is the Rademacher complexity of the family $\mathcal{F}$.
Second, we bound $\mathcal{R}_n$ by
\begin{equation}
    \label{Rademacher}
    \mathcal{R}_n \le 2 ||\w^*||_2  \max_{\x_i \in\mathcal{X}} \sqrt{||x_i||_{\Sinvi}} \sqrt{\frac{2}{n}}
\end{equation}
A special case of this bound was provided by \citet{poulis2017learning} when all samples share the same confidence matrix $\Sinv_i = \Sinv \forall i$ and that matrix is a diagonal matrix with only two allowed values. Here we extend it to the case of multiple confidence matrices. 

Consider first the case where each sample in the data has a confidence matrix that is either $\Sinv_1$ or $\Sinv_2$. For example, this would be the case in a two-class datasets where each class has their own $\Sinv$.
We map each sample $x \in \reals^d$ to a sample $x'\in \reals ^{2d}$ by padding $x$ with $d$ zeros, in a way that depends on its class. If $y=1$ then $x' = [x,\mathbf{0}]$ and if $y=2$ then $x' = [\mathbf{0},x]$. Here, $\mathbf{0}$ is a vector of $d$ zeros. Now define $\Sinv = diag(\Sinv_1,\Sinv_2)$ where $diag$ creates a block diagonal matrix from the two confidence matrices. It is easy to see that 
$|x'|_\Sinv$ equals $|x|_{\Sinv_1}$ when $y=1$ and equals $|x|_{\Sinv_2}$ otherwise.
We also construct a new weight vector $w'\in\reals^{2d}$ by replicating $\w$, such that $w'(d+i) = w(i)$ for any $i\in 1 \ldots d$. It is again easy to see that $\w'^T\x' = \w^T\x$, and that $|w'|_\Sinv = |w|_{\Sinv_1} + |w|_{\Sinv_2}$.

Given this construction, we can now view the data as having a shared confidence matrix. This allows us to apply theorem 5 by \citet{poulis2017learning} to $x'$ and $w'$, this time with a hypothesis family 
$\mathcal{F} = {\w' : ||\w||_\Sinv \le \frac{1}{2}||\w'^*||_\Sinv = \frac{1}{2}||\w^*||_\Sinv}$. This proves \eqref{Rademacher}. The proof of the general case where samples  may have $k$ different confidence matrices $\Sinv_i$ follows the same lines, but replicating $k$ times.

\section{Experiments}
We evaluate the Ellipsotron using two benchmark datasets and compare its performance with two baseline approaches. First, in  a task of scene classification using SUN \citep{xiao2010sun}, a dataset of complex visual scenes accompanied by object segmented by human raters. Second, in a task of recognizing fine-grained classes using CUB \citep{WelinderEtal2010}, a dataset of images of 200 bird species annotated with attributes generated by human raters.

\subsection{Compared Methods}
We tested the following  approaches:

{\bf{(1) Ellipsotron}}. Algorithm 1 described above. 
We used a diagonal matrix $\diinv$, obtaining a value of 1 for relevant features and $\epsilon=10^{-10}$ for irrelevant features. 

{\bf{(2) Lean supervision} (LS)}. No rich supervision signal, linear online classifier with hinge loss trained using standard passive-aggressive \citep{crammer2006online} with  all input features.

{\bf{(3) Feature scaling} (FS)}. Rescale each sample $\x_i$ using its rich supervision matrix $\diinv$, then train with passive-aggressive with the standard hinge loss.
Formally, 
\begin{equation} \label{eq:feature-scaling}
    loss_{FS} = 
    \begin{cases}
        0   \,\, (\wpos -\wneg)^T\diinv\x_i > 1\\    
        1-(\wpos -\wneg)^T\diinv\x_i \,\, \text{ otherwise}
    \end{cases} \quad.
\end{equation}
The update steps are:
\begin{equation} \label{eq:feature-scaling-update}
    \begin{split}
        & \wpos \leftarrow \wpos + \frac{loss_{FS}}{2||\diinv\x_i||^2 + \frac{1}{2C}}{\diinv}x_i \\
        & \wneg \leftarrow \wneg - \frac{loss_{FS}}{2||\diinv\x_i||^2 + \frac{1}{2C}}{\diinv}\x_i \quad.
    \end{split}
\end{equation}
Comparing this loss with the Ellipsotron, \eqref{eq:multiclass-ellipsoid-loss}, reveals two main differences. First, the margin criteria in the FS loss is w.r.t. to the scaled samples $\diinv x_i$, while in the ellipsotron loss, the criteria is that the ellipsoid surrounding $\x_i$ would be correctly classified. Second, when a loss is suffered, the ellipsotron loss is w.r.t. the original sample $\x_i$ while the FS loss is w.r.t. the scaled samples $\diinv x_i$. In the case of ``hard" focuse, namely, setting $\diinv$  to $1$ for relevant features and $0$ for irrelevant features, this is equivalent to zeroing the irrelevant features during learning. In this case weights corresponding to irrelevant features are not updated when a sample is presented. Note that weights for the hardest negative class features usually do not remain at zero, since they experience negative gradients.  

\subsection{Visual Scenes Classification}
The SUN database \citep{xiao2010sun} is a large scale dataset for visual scene recognition.  Each SUN image is accompanied by human-annotated objects visible in the scene and marked with free-text tags like ``table" or ``person", for a total of 4,917 unique tags (see example in Fig. 1). We processed tags by removing suffixes ``\_occluded" and ``\_crop", duplicates (``crop" and ``cropped") and fixing tag typos (``occluded"/ ``occlulded"). The resulting set had 271,737 annotations over a vocabulary of unique 3,772 object tags. Typically, objects tags appear more than once in an image (median over images of object count is 2). We also removed images with encoding issues and images marked as ``misc" and ``outliers", yielding a set of 15,872 images and 1073 scene labels.

\begin{figure}[t]
    \includegraphics[height=2.65cm,trim={0 16.3cm  0 0},clip]{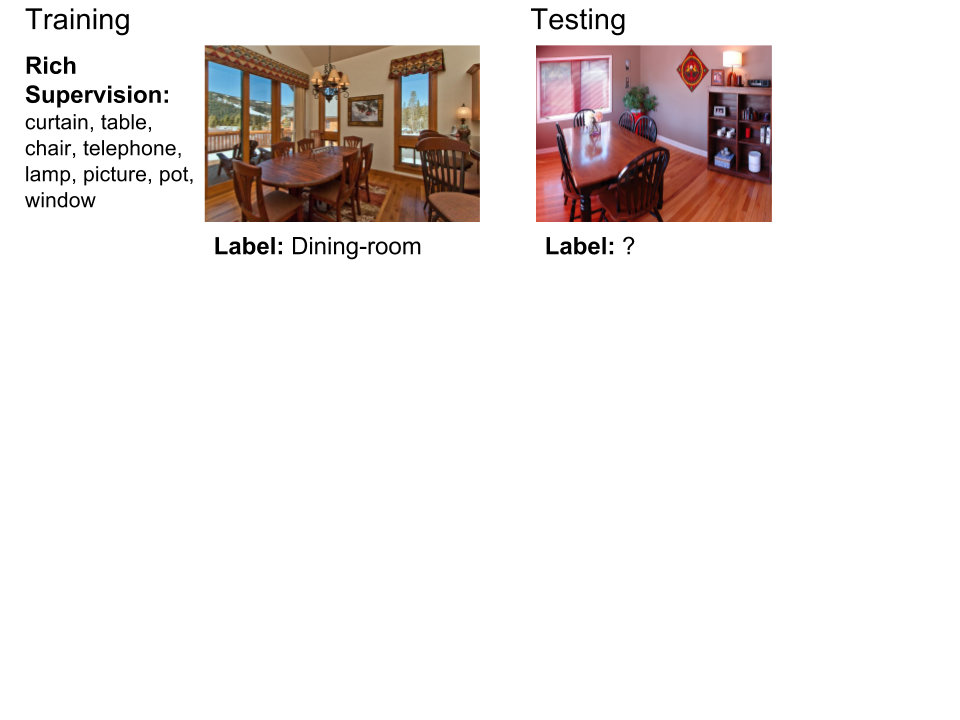} 
    \caption{\textit{An illustration of the learning setup: Labeled samples are accompanied with rich information that provides hints or explains classification. During  training, annotators list objects they observe in a visual scene and tag them with free text (curtain, table, chair, ...). Irrelevant objects in the background are often ignored  (a tree, forest and a ski-slope). At test time, only the image is provided.}}
    \label{fig:illustration}
\end{figure}

Importantly, object tagging in SUN used free-form tags, not restricted to a predefined vocabulary. To test if this type of information can be used as a rich supervision signal, we need an intermediate representation of images that has the following properties: (1) it can be computed on images from new classes without training (since the number of samples per new class is small), and (2) free form tags can be mapped to it, again without training, for the same reason.

\textbf{Representing images with textual terms.}
We mapped images into a vocabulary of 1000 terms using a multilabel classifier based on VGG named {\em visual concepts}  \citep{Fang_2015_CVPR}. Network was originally trained on MS-COCO images and  captions \citep{lin2014microsoft}, yielding a vocabulary that differs from SUN vocabulary of object tags, and contains various types of words present in MS-COCO captions including nouns, adjectives and pronouns.

Importantly, the feature representation was never trained to predict identity of a scene. In this sense, we perform a {\em strong-transfer} from one task (predicting MS-COCO terms) to a different task (scene classification) on a different dataset (SUN). This is different from the more common transfer learning setup where classifiers trained over a subset of classes in a task (say object recognition) are transferred to other classes in the same task (other objects). Strong transfer is a hallmark of high-level abstraction that people exhibit, and is typically hard to achieve.

\textbf{Rich supervision.}
As a source of rich supervision we used the objects detected for each image sample. The intuition is that objects that were marked as present in a scene, can be treated as confidently being detected in the image.
Importantly, providing detected objects is a weak form of rich supervision, because human raters were not instructed to mark objects that are discriminative about the scene or even very relevant to the scene. Indeed, some objects (like people) appear very widely across scenes. 

The set of SUN objects was mapped to VC terms using string matching after stemming both lists. For object tags that contained compound nouns ("TV table"), we matched the VC term with the second term of the phrase (table). Objects that did not match any VC term were removed and not used for rich supervision. With this matching, a total of 1631 SUN objects were matched to 531 VC terms. Rich-supervision is treated as a sample-based binary signal, indicating if an object is present in an image sample. 

\textbf{Evaluation.}
For each scene, we used $50\%$ of samples for testing and the rest for training. Results were hardly sensitive to the value of the complexity hyper parameter $C$ in early experiments, so we fixed its value at $C=1$. Weights were initialized to the zero vector.

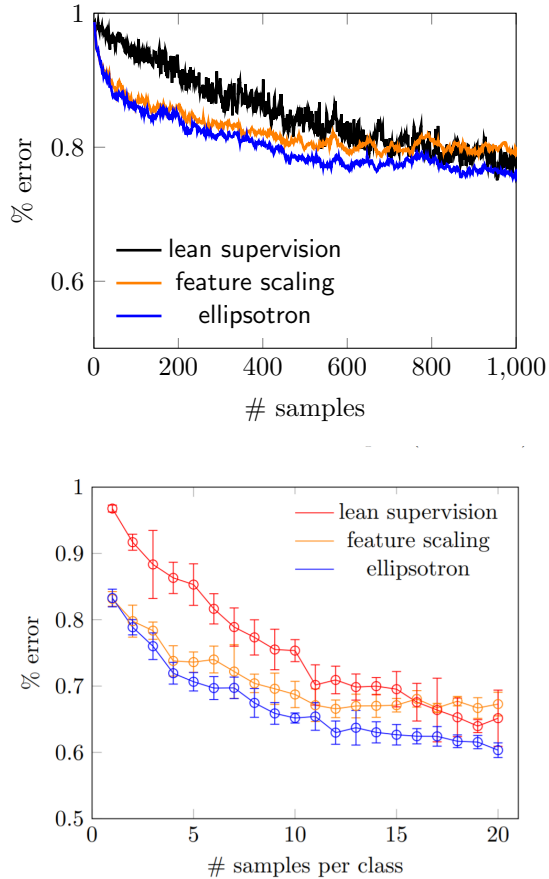
\begin{figure}[ht]
    \begin{center}
        \begin{tikzpicture}
            \begin{axis}
            [height =0.26\textwidth ,width=0.68\linewidth, scale only axis, xlabel = \# samples, xmin = 0, xmax = 1000, ylabel = \% error, ymax = 1.0, ymin = 0.5, ytick pos=left, legend entries={lean supervision, feature scaling, ellipsotron}, legend style={font=\sansmath\sffamily, draw=none, nodes={scale=1.0, transform shape}}, legend pos=south west, mark size=1pt]
            \addplot[clip marker paths=true, line width=1pt, color = black ,mark = None] table [x=steps, y=baselineMean, col sep=comma] {data_files/sun100results.csv};
            \addplot[clip marker paths=true, line width=1pt, color = orange ,mark = None] table [x=steps, y=featuresScalingMean, col sep=comma] {data_files/sun100results.csv};
            \addplot[clip marker paths=true, line width=1pt, solid, color = blue ,mark = None] table [x=steps, y=sampleMean, col sep=comma] {data_files/sun100results.csv};
            \end{axis}
        \end{tikzpicture}
    \end{center}
    \begin{center}
        \includegraphics[width=0.85\linewidth]{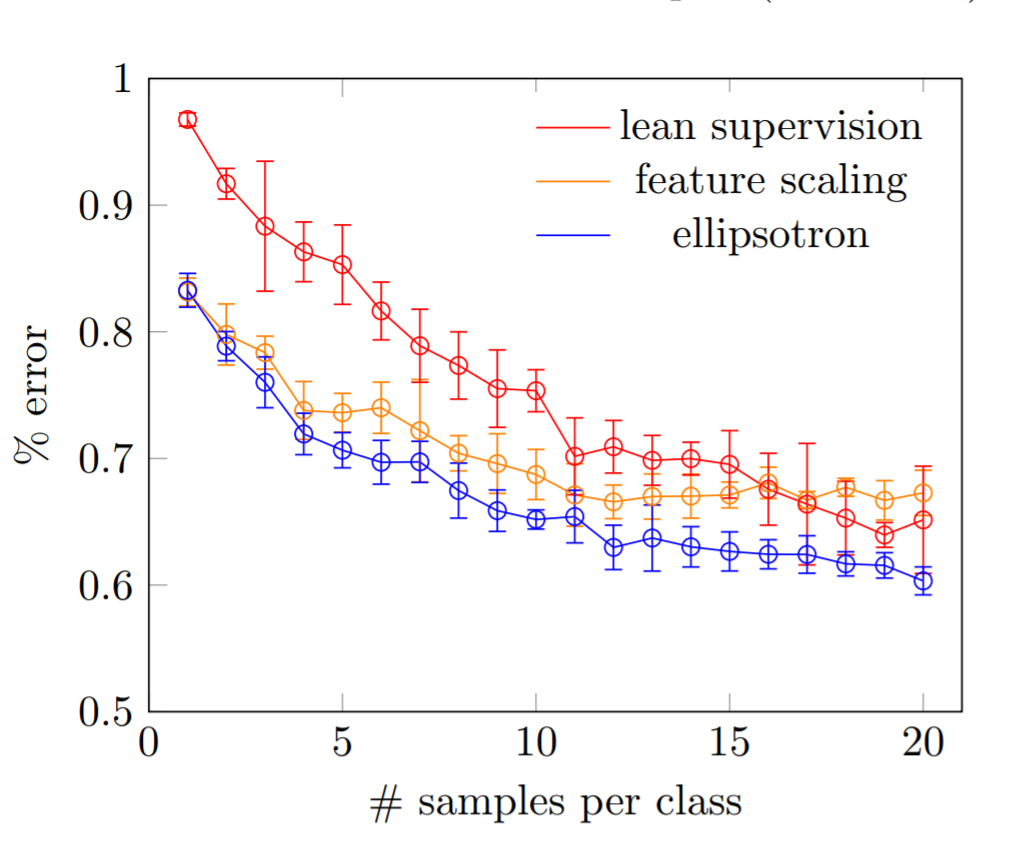}
    \end{center}
    \vspace{-10pt}
    \caption{\textit{{\bf SUN dataset}.
     Mean over 5 random-seed data splits. \textbf{Top:} Test error as a function of training samples observed. 100 classes. \textbf{Bottom:} Test error vs number of samples observed per class. Analyzed classes with 40 to 100 samples (41 classes). Error bars denote the standard error deviation over 5 random-seed data splits.}
    }
    \label{fig:sun41classes_acc}
\end{figure}

\vspace{10pt}
\textbf{Results.}
We first tested Ellipsotron and baselines in a standard online setup where samples are drawn uniformly at random from the training set. Here we used classes that had at least 20 samples, and at most 100 samples, yielding 100 classes. Figure~\ref{fig:sun41classes_acc}(top) shows the accuracy as a function of number of samples (and updates) for SUN data, showing that Ellipsotron outperforms the two baselines when the number of samples is small. 

Classes of visual scenes in SUN differ in terms of the number of samples they have, so averaging across classes unfairly mixes heavy-sampled and poorly-sampled classes. For a more controlled analysis, we analyzed the accuracy as a function of the number of samples per class. Figure~\ref{fig:sun41classes_acc}(bottom) shows the accuracy as a function of training-set size, across 5 randomly drawn training sets of each size. Ellipsotron is consistently more accurate than  baselines for all training-set sizes tested. With 10 samples, the accuracy over the lean  baseline  improves by 33\% (from 25\% to 33\%), and by 10\%  (from 30\% to 33\%), for the feature-scaling baseline.  Table \ref{table:sun_results} provides the cumulative error and cumulative loss, which are common metrics in online-learning. It shows a good agreement between the loss and errors.  %

\subsection{Sensitivity to number of classes}
\begin{figure}[ht]
    \begin{tikzpicture}
        \begin{axis}
        [height =0.3\textwidth ,width=0.8\linewidth, scale only axis, xtick={13, 21, 34, 41, 54, 62, 72, 83,110}, xlabel = \# classes, ylabel = \% error, ymax = 1.0, ymin = 0.0, ytick pos=left, legend entries={lean supervision, feature scaling, ellipsotron}, legend style={font=\sansmath\sffamily, draw=none, nodes={scale=1.0, transform shape}}, mark size=1.5pt, legend pos =south east]
        \addplot[clip marker paths=true, line width=1pt, dashed, color = red ,mark = None, thick] table [x=steps, y=baselineMean, col sep=comma] {data_files/sunCrossClassesResults_5samples.csv};
        \addplot[clip marker paths=true, line width=1pt, dashed, color = orange ,mark = None, thick] table [x=steps, y=hardMean, col sep=comma] {data_files/sunCrossClassesResults_5samples.csv};
        \addplot[clip marker paths=true, line width=1pt, solid, color = blue ,mark = None, thick] table [x=steps, y=sampleMean, col sep=comma] {data_files/sunCrossClassesResults_5samples.csv};
        \addplot[only marks, black, mark=*, mark options={black}, error bars/.cd,y dir=both, y explicit] table [x=steps, y=baselineMean, y error=baselineStd, col sep=comma]{data_files/sunCrossClassesResults_5samples.csv};
        \addplot[only marks, orange, mark=*, mark options={orange}, error bars/.cd,y dir=both, y explicit] table [x=steps, y=hardMean, y error=hardStd, col sep=comma]{data_files/sunCrossClassesResults_5samples.csv};
        \addplot[only marks, blue, mark=*, mark options={blue}, error bars/.cd,y dir=both, y explicit] table [x=steps, y=sampleMean, y error=sampleStd, col sep=comma]{data_files/sunCrossClassesResults_5samples.csv};
        \end{axis}
    \end{tikzpicture}
    \ignore{
    \begin{tikzpicture}
        \begin{semilogxaxis}[log basis x=10,samples at={0.001,...,1000}, width=2.8cm, height=2.3cm, scale only axis, xlabel = \# samples per class, ylabel = \% error, ymax = 1.0, ymin = 0.0, ytick pos=left, legend entries={lean supervision, feature scaling, ellipsotron}, legend style={font=\sansmath\sffamily,draw=none, nodes={scale=0.7, transform shape}}, mark size=1, legend pos =south east ]
        
        \addplot[clip marker paths=true, line width=1pt, dashed, color = red ,mark = None, thin] table [x=steps, y=baselineMean, col sep=comma] {data_files/sun41CscanResults.csv};
        \addplot[clip marker paths=true, line width=1pt, dashed, color = orange ,mark = None, thin] table [x=steps, y=hardMean, col sep=comma] {data_files/sun41CscanResults.csv};
        \addplot[clip marker paths=true, line width=1pt, solid, color = blue ,mark = None, thin] table [x=steps, y=sampleMean, col sep=comma] {data_files/sun41CscanResults.csv};
        \addplot[only marks, orange, mark=*, mark options={orange}, error bars/.cd,y dir=both,  y explicit] table [x=steps, y=hardMean, y error=hardStd, col sep=comma]{data_files/sun41CscanResults.csv};
        \addplot[only marks, black, mark=*, mark options={black}, error bars/.cd,y dir=both,  y explicit] table [x=steps, y=baselineMean, y error=baselineStd, col sep=comma]{data_files/sun41CscanResults.csv};
        \addplot[only marks, blue, mark=*, mark options={blue}, error bars/.cd,y dir=both,  y explicit] table [x=steps, y=sampleMean, y error=sampleStd, col sep=comma]{data_files/sun41CscanResults.csv};
        \end{semilogxaxis}
    \end{tikzpicture}
    }
    \vspace{-10pt}
    \caption{\textit{Percent error on SUN with 5 training samples as a function of the number of classes. Error bars denote the standard error of the mean over 5 repeats.}}
    \label{fig:sun_cross_classes_and_C_scan}
\end{figure}
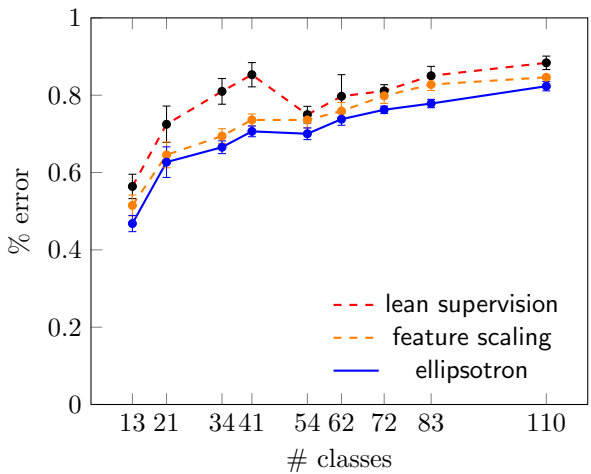

We further repeated the experiments for various number of classes. Figure~\ref{fig:sun_cross_classes_and_C_scan} shows that our results are consistent across various number of classes. Classes were selected by setting upper and lower bonds on the number of samples per class. We varied these bounds between  20 samples and 100 samples. The figure shows the accuracy for training with 5 samples, and similar results are obtained with other number of samples (not shown). 

\begin{table*}[t]
  \begin{center}
    \scalebox{0.99}{
    \begin{tabular}{>{\bfseries}c*{6}{c}}\toprule
    \multirow{2}{*} {\bfseries update steps (samples)} &
    \multicolumn{3}{c}{\bfseries cumulative error \% (avg)} & \multicolumn{3}{c}{\bfseries cumulative loss (avg)} 
    \\\cmidrule(lr){2-4}\cmidrule(lr){5-7}
        & \textbf{5} & \textbf{10} & \textbf{20} & \textbf{5} & \textbf{10} & \textbf{20}   \\ \midrule
        lean            & 89.6 & 83.7 & 75.8 & 1860 & 1803 & 1705  \\
        feature-scaling & 77.73 & 74.36 & 70.74 & 1580 & 1561 & 1545  \\
        ellipsotron     & \textbf{76.14} & \textbf{71.86} & \textbf{67.24} & 1576 & 1566 & 1558  \\
        \midrule
        ellipsotron class-threshold & 75.25 & 71.08 & 66.93 & 1583 & 1581 & 1612  \\
        ellipsotron class-soft   & \textbf{73.13} & \textbf{68.94} & 65.05 & 1370 & 1363 & 1355  \\
        ellipsotron cross-classes   & 78.24 & 72.3 & \textbf{65.04} & 1609 & 1563 & 1492  \\
    \bottomrule
    \end{tabular}
    }
    \caption{\textit{Scene classification cumulative error and cumulative loss on SUN, divided by the number of samples for easier comparison. Top rows are for sample-level RS. Bottom rows are for class-level RS (Sec. \ref{sec_class_level}).}}
  \end{center}
  \label{table:sun_results}
\end{table*}

\subsection{Bird Specie Classification}
As a second set of experiments, we tested Ellipsotron with the CUB database \citep{WelinderEtal2010}. CUB contains 11K images of 200 bird species, where each image is accompanied by attributes like "head is red" and "bill is curved" taken from from a predefined set of 312 attributes. The annotation of attributes per image is done by human non-expert annotators, and was somewhat noisy: attributes are often missing, and sometimes incorrect (the head is orange, not red). We used the attributes as a source of rich supervision for training a bird classifier, on top of a set of attribute predictors. At test time, the classifier maps images to bird classes without the need to specify attributes.

\textbf{Mapping pixels to attributes.}
We used images of 150 species for learning to map pixels to attributes, serving as a representation that rich supervision can interact with. The remaining 50 classes (bird species) were used to evaluate learning with rich supervision. This set contained 2933 images, an average of $\sim 58$ images per class.

To represent each image using the predefined set of attributes, we trained an attribute detector mapping each image onto 312 predefined attributes. The detector is based on resNet50 \citep{he2016deep} trained on ImageNet. We replaced its last, fully-connected, layer with a new fully-connected layer having sigmoid activation on the output. The new layer was  trained with a multilabel binary cross-entropy loss, while keeping the weights of the lower layers frozen. We used 100 bird species drawn randomly to train the attribute predictors, and 50 classes for validation to tune early stopping and hyper parameters. Specifically, we tuned learning rate in [1e-6, 1e-5, 1e-4, 1e-3] and weight decay in [1e-6, 1e-5, 1e-4]. The best training used 1420 steps, 5 epochs, batch size of 32. After selecting hyper parameters using the validation set, models were retrained on all 150 (train and validation) classes.

\textbf{Rich supervision.} We use attributes annotations as a rich-supervision signal. The intuition is that when a rater notes an attribute, it can be treated with more confidence if detected in the image. In practice, attribute annotations vary across same-class images. This happens due to variance in appearance across images in view point and color, or due to different interpretations by different raters. 

\textbf{Experimental setup.} We randomly selected 25 samples of each class for training, and the rest was used as a fixed test set making evaluation more stable. Hyper-parameters were set based on the experiments above with SUN data, setting the aggressiveness at $c=1$. 

\textbf{Results.} Figure (\ref{fig:cub50classes}) depicts percent error as a function of number of training sample on CUB. Ellipsotron consistently improves over lean supervision and feature scaling. With 10 samples, the accuracy over both baselines improves by 44\% (from 18\% to 26\%). Table (\ref{table:cub_results}) shows cumulative accuracy and loss.

\begin{figure}[ht]
    \begin{center}
        \includegraphics[width=0.9\linewidth]{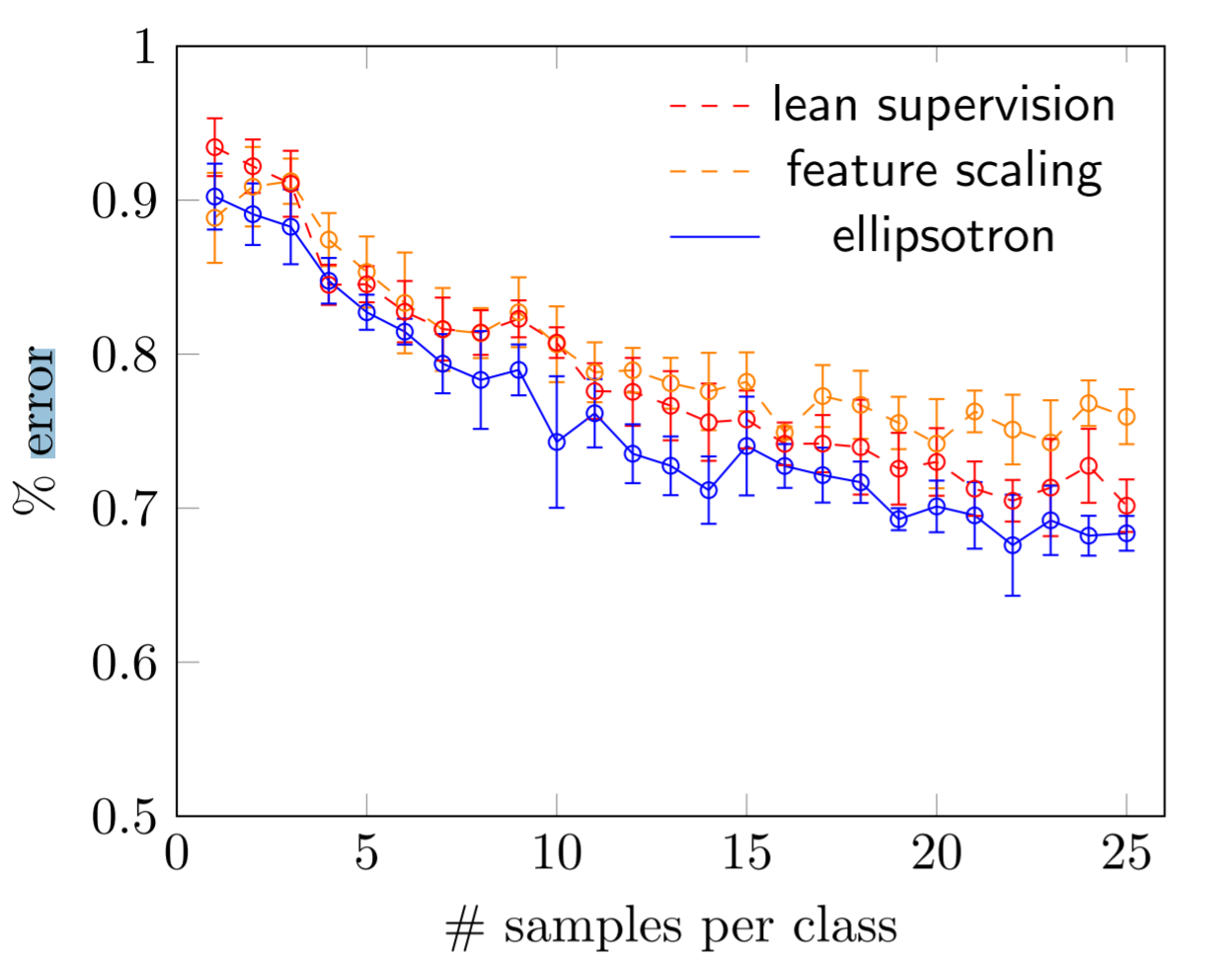}
    \end{center}
    \vspace{-10pt}
    \caption{\textit{CUB bird data. Test error for 50 classes as a function of number of training samples. Error bars denote the standard error of the mean over 5 repeats.}}
    \label{fig:cub50classes}
\end{figure}

\begin{table*}[t]
  \begin{center}
    \scalebox{0.99}{
    \begin{tabular}{>{\bfseries}c*{6}{c}}\toprule
    \multirow{2}{*} {\bfseries update steps (samples)} &
    \multicolumn{3}{c}{\bfseries cumulative error \% (avg)} & \multicolumn{3}{c}{\bfseries cumulative loss (avg)} 
    \\\cmidrule(lr){2-4}\cmidrule(lr){5-7}
        & \textbf{5} & \textbf{10} & \textbf{20} & \textbf{5} & \textbf{10} & \textbf{20}   \\ \midrule
        lean            & 93.44 & 85.45 & 80.28 & 2506 & 2450 & 2364  \\
        feature-scaling & 88.74 & 85.33 & 81.18 & 2523 & 2462 & 2405  \\
        ellipsotron     & \textbf{87} & \textbf{82.75} & \textbf{77.55} & 2570 & 2483 & 2434  \\
        \midrule
        ellipsotron class-threshold & \textbf{86.33} & \textbf{81.38} & 76 & 2482 & 2420 & 2366  \\
        ellipsotron class-soft   & 86.5 & 81.75 & \textbf{75.65} & 1940 & 1917 & 1887  \\
        ellipsotron cross-classes   & 89.15 & 85.45 & 80.3 & 2506 & 2450 & 2364  \\
    \bottomrule
    \end{tabular}
    }
    \caption{\textit{Cumulative error and cumulative loss on CUB. Top rows are for sample-level RS. Bottom rows are for class-level RS (Sec. \ref{sec_class_level}).} }
  \end{center}
  \label{table:cub_results}
\end{table*}

\subsection{Class-Level Supervision}
\label{sec_class_level}
The experiments above tested rich supervision provided at the level of individual samples. However, in some cases, expert can provide information at the level of a class. This is the case when a description of a class is available, or when an expert can provide feedback about which features are important for recognizing a class (``a bathroom should have a sink''). Comparing sample-level supervision and class-level RS is related to the distinction between a {\em  description} of an image, which is sample-dependent but class agnostic, and a {\em class definition} which is sample-agnostic but class dependent. Class-level RS may be able to provide a higher signal-to-noise when it is provided by an expert teacher and shared between all samples of a class, or if it is automatically extracted from aggregating multiple sample-level RS of non-experts. At the same time, it may provide less useful signals if its RS signal does not reflect well uncertainty per example.

Here we evaluated few approaches where class-level RS is aggregated from sample-level RS, aiming to reflect the knowledge that a "teacher" can share with the learning agent.

\subsubsection{Approaches}
We compared three approaches for class-level rich supervision.
For all approaches, the class-level relevance of a feature is computed by aggregating information from sample-level ratings. Specifically, if a class has $n$ samples, then we have $n$ binary votes if a feature $j$ is relevant $a_1,\ldots,a_n \in \{0,1\}$.

{\bf {(1) Class soft}}. We treated the votes as coming from a Poisson distribution and estimated its standard deviation using maximum likelihood. Specifically, $s_i$ is the square root of the fraction of positive votes $s_i = \sqrt{\sum_j a_j/n}$. To avoid a case were some classes have smaller gradients on average than others, we then further normalized the vector of $s_i$ standard deviations to have an $L_2$ norm of 1. 

{\bf {(2) Class threshold}}. We summed the votes, and thresholded, setting $s_i=1$ if $\sum a_j>\Theta$ and $s_i=0$ otherwise. The threshold was selected by training on 10 samples and using the remaining 10 samples as a validation set. We used $\Theta=4$ for SUN experiments and $\Theta=5$ for CUB.

{\bf {(3) Cross classes}}. Using global information about relevant features in all the training data, similar to \citet{poulis2017learning}, $\di$ is global and shared across all training samples. We summed all votes for all classes, setting $s_i=1$ if the feature $i$ it was voted as relevant at least once and $s_i=0$ otherwise.

\begin{figure}[ht]
    \begin{center}
        \includegraphics[width=0.85\linewidth]{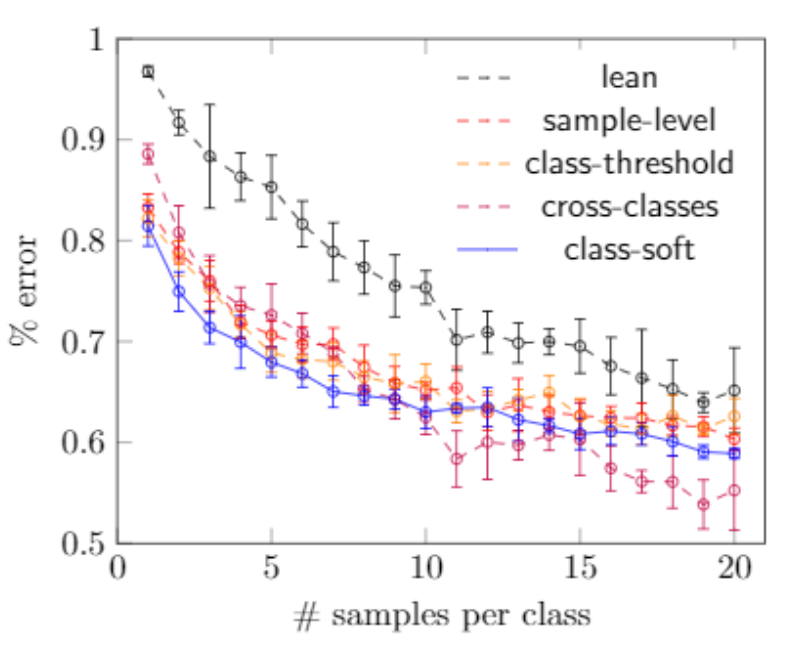}
        \includegraphics[width=0.85\linewidth]{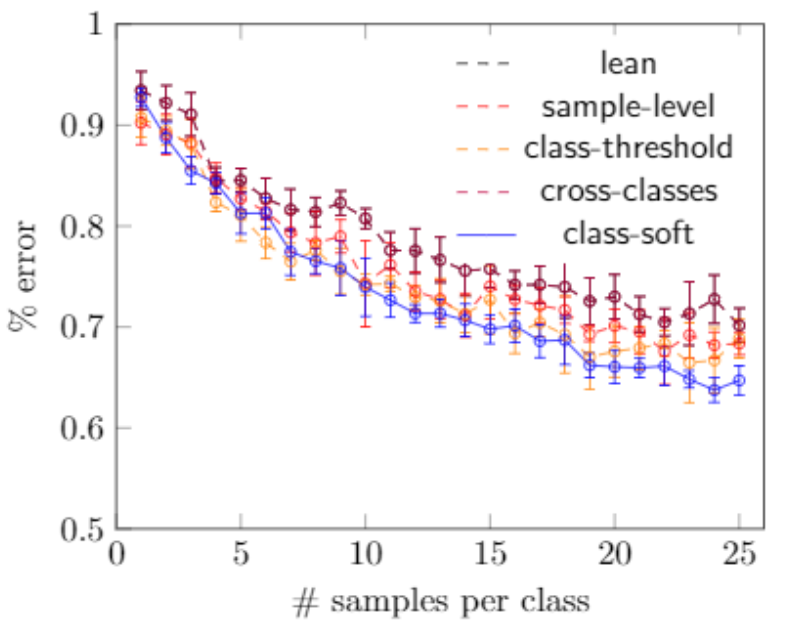}
    \end{center}
    \vspace{-10pt}
    \caption{{\bf Class-based rich supervision}. Percent of test error for all compared methods vs number of training samples, Error bars denote the standard error of the mean over 5 repeats. \textbf{left:} SUN data, classes selected to have sizes between 20 and 100 samples (41 classes). \textbf{Right:} CUB, 50 classes.\newline}
    \label{fig:sun_and_cub_class}
\end{figure}

\subsubsection{Results} We compare class-level and sample-level rich supervising on CUB and SUN datasets descrobed above.

Figure (\ref{fig:sun_and_cub_class})A compares class-level RS with sample-level RS for SUN. Class-soft Ellipsotron outperforms the lean-supervision baseline, the sample-level supervision, as well as class-threshold. With 10 samples, accuracy improves by 44\% (from 25\% to 36\%) over the lean baseline. Cross-classes supervision shows improvement in later stages of learning. Cumulative accuracy and loss are shown in Table (\ref{table:sun_results}).

Figure \ref{fig:sun_and_cub_class}B compares class-level RS  with sample-level and lean supervision on CUB. Class-soft Ellipsotron outperforms all other approaches. With 10 samples, the accuracy  improves by 44\% (from 18\% to 26\%) over the lean baseline. With cross-classes supervision, $\di$ becomes almost identical to the identity matrix and therefore behaves like lean supervision.

\section{Summary}
We presented an online learning approach where labeled samples are also accompanied with rich-supervision. Rich-supervision entails knowledge a teacher has about class features or image features, which in our setup is given in the form of feature uncertainty. The crux of our online approach is to define a sample-dependent margin for each sample, whose multidimensional shape is based on the given information about the uncertainty of features for that particular sample. Experiments on two benchmarks of real-world complex images, for scene classification and fine-grained classification, demonstrate that Ellipsotron outperforms baselines.

\appendix
\section{Proof of Proposition 1}
The algorithm solves the following optimization problem for each sample:
\begin{flalign}
    \label{eq:ellipsotron-optim-sup}
    \min_W & \snormsq{W-W^{t-1}} + C \,\, \loss(W; \cc, y).
\end{flalign}
and the loss was defined as
\begin{equation} \label{eq:multiclass-ellipsoid-loss-proof}
        \loss(W; \cc, y) \!\!= \!\!\\
        \begin{cases}
            0 & \!\!\min\limits_{\around \in \cal{X}_S}\Delta\w^T\around\!>\!0\\    
            1-\Delta\w^T\cc \quad &\text{otherwise}\,.
    \end{cases}
\end{equation}

{\bf Proposition:} {\em The solution to \eqref{eq:ellipsotron-optim-sup} is obtained by the following update steps:}
\begin{equation}
    \begin{split}\label{eq:W_update-sup}
        & {(\wpost)}^T \leftarrow \wposto^T + \frac{\loss}{2||\diinv \x_i||^2 + \frac{1}{2C}} \diinv^T \diinv \x_i \\ 
        & {(\wnegt)}^T \leftarrow \wnegto^T - \frac{\loss}{2||\diinv \x_i||^2 + \frac{1}{2C}} \diinv^T \diinv \x_i \quad.
    \end{split}
\end{equation}

{\bf Proof:}
For a given $\x_i$ with uncertainty matrix $\di$, we use ${\w}^T \x_i = {\w}^T \di \diinv \x_i$ to make a change of variables using $\di$. We denote $\vvec^T = {\w}^T \di$, $\scaled = \diinv \x_i$, so ${\w}^T \x_i$ = $\vvec^T \scaled$. This transformation is applied to the positive class and to the negative class, yielding $\vposto^T = {\wpos}^T \di$,  $\vnegto^T = {\wneg}^T \di$. 

Applying this change of variables to \eqref{eq:multiclass-ellipsoid-loss-proof} yields the following equivalent loss
\begin{equation}
   loss_V(V; \scaled, y_i) \!\!=\!\! 
   \begin{cases}
        0  \,\, \min\limits_{\left\{\uu': ||\uu'\!\!-\!\!\scaled||\le 1/||\Delta\vvec|| \right\}} \, \Delta\vvec^T\uu'>0\\   
        1-\Delta\vvec^T\scaled \,\, \text{otherwise}
\end{cases}
\end{equation}
where $V$ is the weight matrix for all classes, and $\Delta\vvec=\vpos-\vneg$. As done with transforming \eqref{eq:binary-ellipsoid-loss} and since we operate in the transformed "spherized" space, this loss is equivalent to the standard hinge loss $\max(0, 1-(\vpos- \vneg)^T \uu_i)$.

With the change of variables, the optimization problem in \eqref{eq:ellipsotron-optim-sup} becomes equivalent to a standard PA problem:
\begin{flalign}
    \min_V \,\,&||V - V^{t-1}||_{Fro}^2  + C \xi \\ \nonumber
 s.t. \,\,\,\,\
& \vpos^T \scaled - \vneg^T \scaled > 1- \xi \\ \nonumber
& \xi>0
\end{flalign}
whose update steps are \cite[see][]{crammer2006online}
\begin{equation}
    \begin{split}
        & \vpost \leftarrow \vposto + \tau \frac{\partial loss_V}{\partial v} = \vposto + \tau \scaled \\
        & \vnegt \leftarrow \vnegto - \tau \frac{\partial loss_V}{\partial v} = \vnegto - \tau (\scaled) \quad,
    \end{split}
\end{equation}
with $\tau = {loss_V}/\left({2||\scaled||^2 + \frac{1}{2C}}\right)$. Representing the update back in terms of $\w$ and $\x$, %
\begin{equation}
    \begin{split}
        & {(\wpost \di)}^T \leftarrow {(\wposto \di)}^T + \frac{\loss}{2||\diinv \x_i||^2 + \frac{1}{2C}} \diinv \x_i \\
        & {(\wnegt \di)}^T \leftarrow {(\wnegto \di)}^T - \frac{\loss}{2||\diinv \x_i||^2 + \frac{1}{2C}} \diinv \x_i %
    \end{split}
\end{equation}
then multiplying by $\SinviT$ from the left completes the proof.

\ignore{
\section{Figures}
\begin{figure}[ht]
    \begin{tikzpicture}
    \begin{axis}[
    height =0.3\textwidth ,width=0.8\linewidth, scale only axis, xlabel = \# samples per class, xmin = 0, xmax = 21, ylabel = \% error, ymax = 1.0, ymin = 0.5, ytick pos=left, 
    legend entries={lean supervision, feature scaling, ellipsotron}, legend style={draw=none}, mark size=2pt
    ]
    \addplot[clip marker paths=true, line width=1pt, solid, color = red ,mark = None, thin] table [x=steps, y=baselineMean, col sep=comma] {data_files/sun41results.csv};
    \addplot[clip marker paths=true, line width=1pt, solid, color = orange ,mark = None, thin] table [x=steps, y=hardMean, col sep=comma] {data_files/sun41results.csv};
    \addplot[clip marker paths=true, line width=1pt, solid, color = blue ,mark = None, thin] table [x=steps, y=sampleMean, col sep=comma] {data_files/sun41results.csv};
    \addplot[only marks, orange, mark=o, mark options={orange}, error bars/.cd,y dir=both,  y explicit] table [x=steps, y=hardMean, y error=hardStd, col sep=comma]{data_files/sun41results.csv};
    \addplot[only marks, red, mark=o, mark options={red}, error bars/.cd,y dir=both,  y explicit] table [x=steps, y=baselineMean, y error=baselineStd, col sep=comma]{data_files/sun41results.csv};
    \addplot[only marks, blue, mark=o, mark options={blue}, error bars/.cd,y dir=both,  y explicit] table [x=steps, y=sampleMean, y error=sampleStd, col sep=comma]{data_files/sun41results.csv};
    \end{axis}
    \end{tikzpicture}
    \vspace{-10pt}
    \caption{{\bf SUN scene recognition dataset}. Test error for 41 classes as a function of number of training samples. Error bars denote the standard error of the mean over 5 sets drawn with random seeds.  }
    \label{fig:sun41classes_acc}
\end{figure}

\begin{figure}[ht]
    \begin{tikzpicture}
        \begin{axis}[
        height =0.3\textwidth ,width=0.8\linewidth, scale only axis,
        xlabel = \# samples per class, xmin = 0, xmax = 26, ylabel = \% error, ymax = 1.0, ymin = 0.5, ytick pos=left, 
        legend entries={lean supervision, feature scaling, ellipsotron}, legend style={font=\sansmath\sffamily,draw=none}, mark size=1.5pt
        ]
        \addplot[clip marker paths=true, line width=1pt, dashed, color = red ,mark = None, thin] table [x=steps, y=baselineMean, col sep=comma] {data_files/cub50results.csv};
        \addplot[clip marker paths=true, line width=1pt, dashed, color = orange ,mark = None, thin] table [x=steps, y=hardMean, col sep=comma] {data_files/cub50results.csv};
        \addplot[clip marker paths=true, line width=1pt, solid, color = blue ,mark = None, thin] table [x=steps, y=sampleMean, col sep=comma] {data_files/cub50results.csv};
        \addplot[only marks, orange, mark=o, mark options={orange}, error bars/.cd,y dir=both,  y explicit] table [x=steps, y=hardMean, y error=hardStd, col sep=comma]{data_files/cub50results.csv};
        \addplot[only marks, red, mark=o, mark options={red}, error bars/.cd,y dir=both,  y explicit] table [x=steps, y=baselineMean, y error=baselineStd, col sep=comma]{data_files/cub50results.csv};
        \addplot[only marks, blue, mark=o, mark options={blue}, error bars/.cd,y dir=both,  y explicit] table [x=steps, y=sampleMean, y error=sampleStd, col sep=comma]{data_files/cub50results.csv};
        \end{axis}
    \end{tikzpicture}
    \vspace{-10pt}
    \caption{\textit{CUB bird data. Test error for 50 classes as a function of number of training samples. Error bars denote the standard error of the mean over 5 repeats.}}
    \label{fig:cub50classes}
\end{figure}

\begin{figure}[ht]
    \begin{tikzpicture}
    \begin{axis}[
    height =0.3\textwidth ,width=0.8\linewidth, scale only axis,
    xlabel = \# samples per class, xmin = 0, xmax = 21, ylabel = \% error, ymax = 1.0, ymin = 0.5, ytick pos=left, 
    legend entries={lean, sample-level, class-threshold, cross-classes, class-soft}, legend style={font=\sansmath\sffamily,draw=none}, mark size=1.5pt]
    
    \addplot[clip marker paths=true, line width=1pt, dashed, color = black ,mark = None, thin] table [x=steps, y=baselineMean, col sep=comma] {data_files/sun41results.csv};
    \addplot[clip marker paths=true, line width=1pt, dashed, color = red ,mark = None, thin] table [x=steps, y=sampleMean, col sep=comma] {data_files/sun41results.csv};
    \addplot[clip marker paths=true, line width=1pt, dashed, color = orange ,mark = None, thin] table [x=steps, y=thresholdMean, col sep=comma] {data_files/sun41results.csv};
    \addplot[clip marker paths=true, line width=1pt, dashed, color = purple ,mark = None, thin] table [x=steps, y=allinMean, col sep=comma] {data_files/sun41results.csv};
    \addplot[clip marker paths=true, line width=1pt, solid, color = blue ,mark = None, thin] table 
    [x=steps, y=L2scalingMean, col sep=comma] {data_files/sun41results.csv};
    \addplot[only marks, black, mark=o, mark options={black}, error bars/.cd,y dir=both,  y explicit] table [x=steps, y=baselineMean, y error=baselineStd, col sep=comma]{data_files/sun41results.csv};
    \addplot[only marks, red, mark=o, mark options={red}, error bars/.cd,y dir=both,  y explicit] table [x=steps, y=sampleMean, y error=sampleStd, col sep=comma]{data_files/sun41results.csv};
    \addplot[only marks, orange, mark=o, mark options={orange}, error bars/.cd,y dir=both,  y explicit] table [x=steps, y=thresholdMean, y error=thresholdStd, col sep=comma]{data_files/sun41results.csv};
    \addplot[only marks, purple, mark=o, mark options={purple}, error bars/.cd,y dir=both,  y explicit] table [x=steps, y=allinMean, y error=allinStd, col sep=comma]{data_files/sun41results.csv};
    \addplot[only marks, blue, mark=o, mark options={blue}, error bars/.cd,y dir=both,  y explicit] table [x=steps, y=L2scalingMean, y error=L2scalingStd, col sep=comma]{data_files/sun41results.csv};
    \end{axis}
    \end{tikzpicture}
    \vspace{-10pt}
    \caption{\textit{{\bf Class-based rich supervision on SUN}. Percent of test error for all compared methods vs number of training samples, for 41 classes. Error bars denote the standard error of the mean over 5 repeats. }}
    \label{fig:sun_class_vs_sample_41classes}
\end{figure}

\begin{figure}[ht]
    \begin{tikzpicture}
    \begin{axis}[
    height =0.3\textwidth ,width=0.8\linewidth, scale only axis,
    xlabel = \# samples per class, xmin = 0, xmax = 26, ylabel = \% error, ymax = 1.0, ymin = 0.5, ytick pos=left, 
    legend entries={lean, sample-level, class-threshold, cross-classes, class-soft}, legend style={font=\sansmath\sffamily,draw=none}, mark size=1.5pt]
    
    \addplot[clip marker paths=true, line width=1pt, dashed, color = black ,mark = None, thin] table [x=steps, y=baselineMean, col sep=comma] {data_files/cub50results.csv};
    \addplot[clip marker paths=true, line width=1pt, dashed, color = red ,mark = None, thin] table [x=steps, y=sampleMean, col sep=comma] {data_files/cub50results.csv};
    \addplot[clip marker paths=true, line width=1pt, dashed, color = orange ,mark = None, thin] table [x=steps, y=thresholdMean, col sep=comma] {data_files/cub50results.csv};
    \addplot[clip marker paths=true, line width=1pt, dashed, color = purple ,mark = None, thin] table [x=steps, y=allinMean, col sep=comma] {data_files/cub50results.csv};
    \addplot[clip marker paths=true, line width=1pt, solid, color = blue ,mark = None, thin] table [x=steps, y=L2scalingMean, col sep=comma] {data_files/cub50results.csv};
    \addplot[only marks, black, mark=o, mark options={black}, error bars/.cd,y dir=both,  y explicit] table [x=steps, y=baselineMean, y error=baselineStd, col sep=comma]{data_files/cub50results.csv};
    \addplot[only marks, red, mark=o, mark options={red}, error bars/.cd,y dir=both,  y explicit] table [x=steps, y=sampleMean, y error=sampleStd, col sep=comma]{data_files/cub50results.csv};
    \addplot[only marks, orange, mark=o, mark options={orange}, error bars/.cd,y dir=both,  y explicit] table [x=steps, y=thresholdMean, y error=thresholdStd, col sep=comma]{data_files/cub50results.csv};
    \addplot[only marks, purple, mark=o, mark options={purple}, error bars/.cd,y dir=both,  y explicit] table [x=steps, y=allinMean, y error=allinStd, col sep=comma]{data_files/cub50results.csv};
    \addplot[only marks, blue, mark=o, mark options={blue}, error bars/.cd,y dir=both,  y explicit] table [x=steps, y=L2scalingMean, y error=L2scalingStd, col sep=comma]{data_files/cub50results.csv};
    \end{axis}
    \end{tikzpicture}
    \vspace{-10pt}
    \caption{\textit{\textbf{Class-based rich supervision on 50 CUB classes}. Percent of test error for compared methods as a function of number of training samples. Error bars denote the standard error of the mean over 5 repeats.}}
    \label{fig:cub_class_vs_sample_50classes}
\end{figure}

}
\bibliography{ellipsotron}

\begin{thebibliography}{}

\bibitem[Atzmon and Chechik, 2018]{LAGO}
Atzmon, Y. and Chechik, G. (2018).
\newblock Probabilistic and-or attribute grouping for zero-shot learning.
\newblock In {\em UAI}.

\bibitem[Atzmon and Chechik, 2019]{COSMO}
Atzmon, Y. and Chechik, G. (2019).
\newblock Adaptive confidence smoothing for generalized zero-shot learning.
\newblock In {\em CVPR}.

\bibitem[Branson et~al., 2010]{branson2010visual}
Branson, S., Wah, C., Schroff, F., Babenko, B., Welinder, P., Perona, P., and
  Belongie, S. (2010).
\newblock Visual recognition with humans in the loop.
\newblock {\em ECCV 2010}, pages 438--451.

\bibitem[Chechik et~al., 2007]{chechik2007max}
Chechik, G., Heitz, G., Elidan, G., Abbeel, P., and Koller, D. (2007).
\newblock Max-margin classification of incomplete data.
\newblock In {\em NIPS}, pages 233--240.

\bibitem[Crammer et~al., 2006]{crammer2006online}
Crammer, K., Dekel, O., Keshet, J., Shalev-Shwartz, S., and Singer, Y. (2006).
\newblock Online passive-aggressive algorithms.
\newblock {\em J. Machine Learning Research}, 7(Mar):551--585.

\bibitem[Dasgupta et~al., 2018]{Dasgupta_NIPS2018_7651}
Dasgupta, S., Dey, A., Roberts, N., and Sabato, S. (2018).
\newblock Learning from discriminative feature feedback.
\newblock In Bengio, S., Wallach, H., Larochelle, H., Grauman, K.,
  Cesa-Bianchi, N., and Garnett, R., editors, {\em Advances in Neural
  Information Processing Systems 31}, pages 3955--3963. Curran Associates, Inc.

\bibitem[Druck et~al., 2007]{druck2007reducing}
Druck, G., Mann, G., and McCallum, A. (2007).
\newblock Reducing annotation effort using generalized expectation criteria.
\newblock Technical report, Mass. Univ Amherst, CS Dept.

\bibitem[Elhoseiny et~al., 2017]{elhoseiny2017link}
Elhoseiny, M., Zhu, Y., Zhang, H., and Elgammal, A. (2017).
\newblock Link the head to the" beak": Zero shot learning from noisy text
  description at part precision.
\newblock In {\em 2017 IEEE Conference on Computer Vision and Pattern
  Recognition (CVPR)}, pages 6288--6297. IEEE.

\bibitem[Fang et~al., 2015]{Fang_2015_CVPR}
Fang, H., Gupta, S., Iandola, F., Srivastava, R.~K., Deng, L., Dollar, P., Gao,
  J., He, X., Mitchell, M., Platt, J., Zitnick, L., and Zweig, G. (2015).
\newblock From captions to visual concepts and back.
\newblock In {\em CVPR}.

\bibitem[Finn et~al., 2017]{finn2017model}
Finn, C., Abbeel, P., and Levine, S. (2017).
\newblock Model-agnostic meta-learning for fast adaptation of deep networks.
\newblock {\em arXiv preprint arXiv:1703.03400}.

\bibitem[Hariharan and Girshick, 2016]{hariharan2016low}
Hariharan, B. and Girshick, R. (2016).
\newblock Low-shot visual object recognition.
\newblock {\em arXiv preprint arXiv:1606.02819}.

\bibitem[He et~al., 2016]{he2016deep}
He, K., Zhang, X., Ren, S., and Sun, J. (2016).
\newblock Deep residual learning for image recognition.
\newblock In {\em CVPR}, pages 770--778.

\bibitem[Lin et~al., 2014]{lin2014microsoft}
Lin, T.-Y., Maire, M., Belongie, S., Hays, J., Perona, P., Ramanan, D.,
  Doll{\'a}r, P., and Zitnick, L. (2014).
\newblock Microsoft coco: Common objects in context.
\newblock In {\em ECCV}, pages 740--755. Springer.

\bibitem[Mac~Aodha et~al., 2018]{Aodha_2018_CVPR}
Mac~Aodha, O., Su, S., Chen, Y., Perona, P., and Yue, Y. (2018).
\newblock Teaching categories to human learners with visual explanations.
\newblock In {\em The IEEE Conference on Computer Vision and Pattern
  Recognition (CVPR)}.

\bibitem[Poulis and Dasgupta, 2017]{poulis2017learning}
Poulis, S. and Dasgupta, S. (2017).
\newblock Learning with feature feedback: from theory to practice.
\newblock In {\em Artificial Intelligence and Statistics}, pages 1104--1113.

\bibitem[Raghavan et~al., 2006]{raghavan2006active}
Raghavan, H., Madani, O., and Jones, R. (2006).
\newblock Active learning with feedback on features and instances.
\newblock {\em J. Machine Learning Research}, 7:1655--1686.

\bibitem[Ravi and Larochelle, 2017]{ravi2016optimization}
Ravi, S. and Larochelle, H. (2017).
\newblock Optimization as a model for few-shot learning.
\newblock In {\em ICLR}.

\bibitem[Shalev-Shwartz and Singer, 2005]{shalev2005new}
Shalev-Shwartz, S. and Singer, Y. (2005).
\newblock A new perspective on an old perceptron algorithm.
\newblock In {\em COLT}, pages 264--278. Springer.

\bibitem[Small et~al., 2011]{small2011constrained}
Small, K., Wallace, B., Trikalinos, T., and Brodley, C.~E. (2011).
\newblock The constrained weight space svm: learning with ranked features.
\newblock In {\em ICML}, pages 865--872.

\bibitem[Snell et~al., 2017]{snell2017prototypical}
Snell, J., Swersky, K., and Zemel, R.~S. (2017).
\newblock Prototypical networks for few-shot learning.
\newblock {\em arXiv preprint arXiv:1703.05175}.

\bibitem[Su et~al., 2017]{su2017interpretable}
Su, S., Chen, Y., Mac~Aodha, O., Perona, P., and Yue, Y. (2017).
\newblock Interpretable machine teaching via feature feedback.

\bibitem[Sun and DeJong, 2005]{sun2005explanation}
Sun, Q. and DeJong, G. (2005).
\newblock Explanation-augmented svm: an approach to incorporating domain
  knowledge into svm learning.
\newblock In {\em ICML)}, pages 864--871.

\bibitem[Sun et~al., 2006]{sun2006explanation}
Sun, Q., Wang, L.-L., and DeJong, G. (2006).
\newblock Explanation-based learning for image understanding.
\newblock In {\em Proc. Nat. Conf. on Artificial Intelligence}, volume~21, page
  1679.

\bibitem[Sun et~al., 2007]{sun2007robustness}
Sun, Q., Wang, L.-L., Lim, S.~H., and DeJong, G. (2007).
\newblock Robustness through prior knowledge: using explanation-based learning
  to distinguish handwritten chinese characters.
\newblock {\em Int. J. on Document Analysis and Recognition}, 10(3):175--186.

\bibitem[Thrun, 2012]{thrun2012explanation}
Thrun, S. (2012).
\newblock {\em Explanation-based neural network learning: A lifelong learning
  approach}, volume 357.
\newblock Springer Science \& Business Media.

\bibitem[Vinyals et~al., 2016]{vinyals2016matching}
Vinyals, O., Blundell, C., Lillicrap, T., Wierstra, D., et~al. (2016).
\newblock Matching networks for one shot learning.
\newblock In {\em NIPS}, pages 3630--3638.

\bibitem[Welinder et~al., 2010]{WelinderEtal2010}
Welinder, P., Branson, S., Mita, T., Wah, C., Schroff, F., Belongie, S., and
  Perona, P. (2010).
\newblock {Caltech-UCSD Birds 200}.
\newblock Technical Report CNS-TR-2010-001, CalTech.

\bibitem[Xian et~al., 2017]{xian2017zero}
Xian, Y., Schiele, B., and Akata, Z. (2017).
\newblock Zero-shot learning-the good, the bad and the ugly.
\newblock {\em arXiv preprint arXiv:1703.04394}.

\bibitem[Xiao et~al., 2010]{xiao2010sun}
Xiao, J., Hays, J., Ehinger, K.~A., Oliva, A., and Torralba, A. (2010).
\newblock Sun database: Large-scale scene recognition from abbey to zoo.
\newblock In {\em CVPR}, pages 3485--3492.

\bibitem[Zaidan et~al., 2007]{zaidan2007using}
Zaidan, O., Eisner, J., and Piatko, C.~D. (2007).
\newblock Using" annotator rationales" to improve machine learning for text
  categorization.
\newblock In {\em HLT-NAACL}, pages 260--267.

\end{thebibliography}

\end{document}